\documentclass[letterpaper,10pt,conference]{ieeeconf}

\IEEEoverridecommandlockouts  
                            
\usepackage{graphicx}
\usepackage{balance}
\usepackage{comment}
\usepackage{cite}
\usepackage{amssymb}
\usepackage[tight,footnotesize]{subfigure}
\usepackage[active]{srcltx}
\usepackage{amsmath}
\usepackage{mathtools}

\graphicspath{{./figs/}}
\renewcommand{\baselinestretch}{1}
\usepackage{eurosym}
\usepackage[plain]{algorithm}
\usepackage{algorithmic}
\usepackage{multicol}
\usepackage{dsfont}
\usepackage{amsfonts}

\usepackage{cases}
\usepackage{xcolor}
\usepackage{mathrsfs}
\newtheorem{definition}{Definition}
\newtheorem{lemma}{Lemma}

\begin{document}

\title{\LARGE \bf
Data-Driven Predictive Planning and Control for \\Aerial 3D Inspection with Back-face Elimination
}

\author{Savvas~Papaioannou,~Panayiotis~Kolios,~Christos~G.~Panayiotou and~Marios~M.~Polycarpou
\thanks{The authors are with the KIOS Research and Innovation Centre of Excellence (KIOS CoE) and the Department of Electrical and Computer Engineering, University of Cyprus, Nicosia, 1678, Cyprus. E-mail:{\tt\small \{papaioannou.savvas, pkolios, christosp, mpolycar\}@ucy.ac.cy}. %
This work is implemented under the Border Management and Visa Policy Instrument (BMVI) and is co-financed by the European Union and the Republic of Cyprus (GA:BMVI/2021-2022/SA/1.2.1/015), and supported by the European Union's Horizon 2020 research and innovation programme under grant agreement No 739551 (KIOS CoE), and through the Cyprus Deputy Ministry of Research, Innovation and Digital Policy of the Republic of Cyprus.}}

\maketitle

\begin{abstract}

Automated inspection with Unmanned Aerial Systems (UASs) is a transformative capability set to revolutionize various application domains. However, this task is inherently complex, as it demands the seamless integration of perception, planning, and control which existing approaches often treat separately. Moreover, it requires accurate long-horizon planning to predict action sequences, in contrast to many current techniques, which tend to be myopic. 
To overcome these limitations, we propose a 3D inspection approach that unifies perception, planning, and control within a single data-driven predictive control framework. Unlike traditional methods that rely on known UAS dynamic models, our approach requires only input-output data, making it easily applicable to off-the-shelf black-box UASs. Our method incorporates back-face elimination, a visibility determination technique from 3D computer graphics, directly into the control loop, thereby enabling the online generation of accurate, long-horizon 3D inspection trajectories.

\end{abstract}

\section{Introduction} \label{sec:Introduction}

Unmanned Aerial Systems (UAS) are increasingly employed across a range of application domains, such as reconnaissance  tasks \cite{valianti2020multi,papaioannou2023joint}, search-and-rescue operations \cite{moon2021deep,papaioannou2020coordinated,papaioannou2024synergising}, and inspection missions \cite{Zacharia2023,papaioannou2025rolling}.
Despite progress in automation and aviation, fully autonomous UASs with integrated planning and control remain in their infancy. Automated robotic inspection tasks require precise navigation, efficient trajectory planning, and accurate path execution. Long-horizon planning is critical for predicting future actions and optimizing cumulative rewards. However, many existing approaches rely on short-horizon planning in simplified 2D environments, neglecting the complexity of 3D settings \cite{huang2017review,kan2020online}. Conventional methods assume known UAS dynamics, either through explicit modeling \cite{Lindqvist2020,Guevara2024} or learning-based approaches \cite{Saviolo2022,Bauersfeld2021NeuroBEMHA}. Model-based techniques often lead to intractable non-linear optimization problems, while learning-based methods demand large datasets and retraining when applied to new platforms. Recent efforts favor data-driven approaches \cite{Markovsky2008data,Coulson2019data}, which leverage past input-output data without requiring explicit models. 

Moreover, automated inspection is inherently a challenging task because it requires tight integration of perception, planning, control, whereas in many other planning problems these tasks are often handled separately \cite{Xing2023, Goricanec2023, heng2011autonomous,papaioannou2024hierarchical,droeschel2016multilayered}. For instance, \cite{Xing2023, Goricanec2023} address power-line inspection and trajectory following, respectively, assuming pre-defined reference trajectories, effectively treating the planning problem as solved. In \cite{heng2011autonomous}, planning is decoupled into path searching and motion control, while the works in \cite{droeschel2016multilayered,papaioannou2024hierarchical} treat perception, planning, and control as separate modules for UAV navigation and coverage. In contrast, automated inspection demands the joint solution of planning and control under perception-aware constraints. The UAS must dynamically determine the inspection sequence, timing, and reference trajectory to cover multiple surface points, significantly increasing problem complexity.

To address these challenges, this work formulates the 3D inspection task as an optimization problem that maximizes cumulative reward over a receding planning horizon, integrating perception, planning, and control within a data-driven predictive control framework. The proposed method jointly optimizes the UAS motion control inputs and camera viewpoints, incorporating back-face elimination, a technique from 3D computer graphics used for visibility determination, directly into the control loop. This enables the generation of accurate, long-horizon 3D inspection trajectories.

The rest of the paper is organized as follows: Section~\ref{sec:Related_Work} reviews related work. Section~\ref{sec:system_model} presents the problem formulation. Section~\ref{sec:approach} describes the proposed approach, while Section~\ref{sec:Evaluation} provides the evaluation. Finally, Section~\ref{sec:conclusion} concludes the paper.
\section{Related Work}\label{sec:Related_Work}

Most existing inspection planning approaches primarily address 2D environments, such as terrain coverage \cite{tan2021comprehensive}. Visibility-aware methods with fixed sensors, as presented in \cite{kantaros2015distributed}, rely on myopic strategies, limiting their capacity for long-horizon planning. In 3D environments, \cite{bircher2018receding} formulates the inspection task as a sampling-based path planning problem, enabling online receding horizon planning through geometric random trees. The method in \cite{jing2017model} decomposes the problem into view and motion planning, solved heuristically via genetic algorithms. A finite element method (FEM)-based approach is introduced in \cite{ivic2023multi}, where offline inspection paths are generated using potential and distance fields. Particle swarm optimization (PSO) is employed in \cite{shang2020co} and \cite{phung2017enhanced}, with the latter framing the problem as discrete graph optimization. Recent works \cite{shi2021inspection,icuas2024} formulate the planning problem as a traveling salesman problem (TSP), solved using meta-heuristic techniques. Receding horizon control has also been applied to 3D inspection planning in \cite{papaioannou2022uav,papaioannou2023distributed}, however, these methods are constrained to cuboid-like structures and do not incorporate visibility considerations. Finally, the approach in \cite{Zacharia2023} is limited to the generation of myopic plans.
\section{Problem Formulation} \label{sec:system_model}

\subsection{UAS Dynamical Model} \label{ssec:kinematic_model}

Without loss of generality, we assume that the dynamical behavior $\mathscr{B}$ of a UAS agent can be described by a linear time-invariant (LTI) system of the following form \cite{how2015linear}:
\begin{equation} \label{eq:LTI}
\mathscr{B}(A,B,C,D) := 
\begin{cases}
x(t + 1) = Ax(t) + Bu(t) \\
y(t) = Cx(t) + Du(t),
\end{cases}
\end{equation}
where $\mathscr{B}(A,B,C,D)$ is a minimal input/output/state representation of the system, with \( A \in \mathbb{R}^{n \times n} \), \( B \in \mathbb{R}^{n \times m} \), \( C \in \mathbb{R}^{p \times n} \), and \( D \in \mathbb{R}^{p \times m} \)  unknown. The state, control input, and output of the system at time-step \(t \in \mathbb{N}\) are given respectively by \( x(t) \in \mathbb{R}^n \), \( u(t) \in \mathbb{R}^m \),  and \( y(t) \in \mathbb{R}^p \). 
\begin{definition}
The signal $u \in \mathbb{R}^{m \times T}$ of length $T \in \mathbb{N}^{+}$ with $u(t) \in \mathbb{R}^m, t\in\{1,\ldots,T\}$ is persistently exciting of order $N \in \mathbb{N}^{+} \leq T$ if the Hankel matrix:
\begin{equation}
\mathscr{H}_N(u) := 
\begin{pmatrix}
u(1) & u(2) & \dots & u(T-N+1) \\
u(2) & u(3) & \dots & u(T-N+2) \\
\vdots & \vdots & \ddots & \vdots \\
u(N) & u(N+1) & \dots & u(T)
\end{pmatrix},
\end{equation}
has full rank, i.e., $\text{rank}(\mathscr{H}_N(u)) = mN$.
\end{definition}

\begin{lemma}[Fundamental Lemma\cite{Markovsky2008data}]

Consider a controllable LTI system $\mathscr{B}(A,B,C,D)$ described by Eq. \eqref{eq:LTI} and assume that the input sequence (i.e., control input samples) $u^d \in \mathbb{R}^{m \times T}$ is persistently exciting of order  $N+n$, and $y^d \in \mathbb{R}^{p \times T}$ is the corresponding output (i.e., collected output data samples). Then, the input-output pair of sequences $u$ and $y$ of length $N$ is a valid input-output trajectory of $\mathscr{B}(A,B,C,D)$ if and only if there exists $g \in \mathbb{R}^{T-N+1}$ such that:
\begin{equation}
\begin{bmatrix}
\mathscr{H}_N(u^d) \\
\mathscr{H}_N(y^d)
\end{bmatrix} g =
\begin{bmatrix}
u \\
y 
\end{bmatrix}.
\end{equation}
Subsequently, a valid input-output trajectory \( (u, y) \) of the system of length \( N \) can be constructed by collecting a sufficiently rich and long  (i.e., persistently exciting) input-output sequence \( (u^d, y^d) \) of length \( T \) such that \( T \geq (m + 1)(N + n) - 1 \). In this case, the subspace consisting of all valid input and output trajectories of length \( N \) of the system is the same as the range space of the Hankel matrices \( \mathscr{H}_N(u^d) \) and \( \mathscr{H}_N(y^d) \) respectively. For extensions to nonlinear and stochastic systems, we refer readers to \cite{Huang2024}.
\end{lemma}

\subsection{UAS Camera Model} \label{ssec:cam_model}

The UAS carries a gimballed camera module with a finite field of view (FOV), which is utilized for performing automated inspection missions. The camera FOV is represented as a convex polyhedron i.e., a regular right pyramid with square base, characterized by its vertices $V_\text{FOV} \in \mathbb{R}^{3 \times 5}$, where $V_\text{FOV}(i) \in \mathbb{R}^3, i \in \{1,\ldots,5\}$. $V_\text{FOV}$ is adjusted in three-dimensional space by instructing the gimbal controller to perform two sequential basic rotations: first, a rotation by an angle \(\theta_y \in \Theta_y\) around the \(y\)-axis, and then a rotation by an angle \(\theta_z \in \Theta_z\) around the \(z\)-axis as:
\begin{equation}\label{eq:fov_eq}
   V_{\theta_z,\theta_y}(i) = R_{z}(\theta_z) R_{y}(\theta_y) V_\text{FOV}(i) , i\in \{1,\ldots,5\},
\end{equation}
\noindent where, with slight abuse of notation, \( V_{\theta_z,\theta_y} \) is the matrix containing the rotated vertices of the FOV corresponding to the angles \( (\theta_z, \theta_y) \), which are used here as indices. \( R_{y}(\theta) \) and \( R_{z}(\theta) \) are the basic \( 3 \times 3 \) rotation matrices that rotate vectors by an angle \( \theta \) around the \( y \)-axis and \( z \)-axis, respectively. 

The convex hull of $V_{\theta_z,\theta_y}$ is defined as $C_{\theta_z,\theta_y} = \{\text{x} \in \mathbb{R}^3 | \Gamma_{\theta_z,\theta_y} \text{x} \leq \Delta_{\theta_z,\theta_y}\}$, where $\Gamma_{\theta_z,\theta_y}$ is a $5 \times 3$ matrix whose rows are the outward normal vectors $\Gamma^\top_{\theta_z,\theta_y}(i) \in \mathbb{R}^3, i\in\{1,\ldots,5\}$ on the pyramid FOV facets formed by the vertices in $V_{\theta_z,\theta_y}$ and $\Delta_{\theta_z,\theta_y}$ is a $5 \times 1$ constant vector containing the offset $\Delta_{\theta_z,\theta_y}(i)$ of each facet $i$ from the origin. Finally, the set of points $C_{\theta_z,\theta_y}(t)$ that can be observed through the FOV $V_{\theta_z,\theta_y}$ along the agent's trajectory described by Eq. \eqref{eq:LTI} at time-step $t$ is given by:
\begin{equation}
	C_{\theta_z,\theta_y}(t) = \{\text{x} \in \mathbb{R}^3 | \Gamma_{\theta_z,\theta_y} \text{x} \leq \Delta_{\theta_z,\theta_y} + \Gamma_{\theta_z,\theta_y} hy(t)\},
\end{equation}
\noindent where the output of the system $hy(t) \in \mathbb{R}^3$ denotes the UAS position (3D Cartesian coordinates) at time-step $t$, i.e., the matrix $h$ extracts the spatial coordinates from the measurement vector $y(t)$.

\subsection{Problem Statement}\label{ssec:problem}

Consider a known structure of interest, denoted by $\mathcal{S} \subset \mathbb{R}^3$, whose surface is represented by a mesh $\Delta \mathcal{S}$ composed of triangular facets. Let  $\Delta \hat{\mathcal{S}}=\{s_1, \ldots, s_{|\Delta \hat{\mathcal{S}|}}\} \subseteq \Delta \mathcal{S}$ be a subset of facets. Each facet $s_i \in \Delta \hat{\mathcal{S}}$ is defined by three vertices, $\nu s^i_1$, $\nu s^i_2$, and $\nu s^i_3 \in \mathbb{R}^3$. The centroid of each facet is given by $\bar{s}_i = \text{mean}(\nu s^i_1, \nu s^i_2, \nu s^i_3)$, and each facet is associated with an outward normal vector, denoted as $\vec{s}_i$, and a reward value, $r_i$ which may vary across facets.

\textit{The objective of the UAS agent is to design and execute inspection trajectories over a rolling planning horizon of \( N \) time-steps, aiming to maximize the cumulative reward obtained by inspecting the facets \( s_i \in \Delta \hat{\mathcal{S}} \) with its camera, and to conclude the mission once all facets in \( \Delta \hat{\mathcal{S}} \) have been inspected}. 

Let $\gamma_i$ denote a decision variable indicating whether facet $s_i$ has been inspected by the end of a planning horizon of length $N$  time-steps. The agent's objective can be formulated as: $\max \sum_{i=1}^{|\Delta \hat{\mathcal{S}}|} r_i \gamma_i(N)$. In the scenario where $r_i=r, \forall i \in \{1,\ldots,|\Delta \hat{\mathcal{S}}|\}$ the objective simplifies to maximizing the number of facets inspected within the planning horizon. On the other hand, by assigning different reward values to the facets, we can design trajectories that prioritize different parts of the structure to be inspected. 
However, the objective above does not capture the requirement to inspect all facets in  $\Delta \hat{\mathcal{S}}$. To accomplish this, the agent must track which facets have been inspected, ensuring mission progress and avoiding redundant work. Additionally, the agent must determine in an on-line fashion the visibility of each facet based on its predicted state, enabling the planning of trajectories that ensure the inspection of visible areas on the structure's surface.

In summary, the agent is required to inspect all facets, collecting the reward from each facet only once and only if the facet is visible. Next, we discuss how inspection planning dynamics and perception-aware constraints via back-face elimination have been integrated within a data-driven predictive control framework to tackle this problem.

\section{Planning and Control for 3D Inspection}\label{sec:approach}

\subsection{Inspection Planning Dynamics} \label{ssec:inspection_dynamics}

As previously mentioned, the task is to inspect $|\Delta \hat{\mathcal{S}}|$ facets on the surface of $\mathcal{S}$. The inspection value of each facet $s_i$ over the planning horizon is represented by the binary decision variable $\gamma_i(t+\tau|t) \in \{0,1\}$, for $\tau \in \{0,\ldots,N-1\}$. Here, $\gamma_i(t+\tau|t) = 1$ indicates that facet $s_i$ is predicted to be inspected at time-step $t+\tau$ of the planning horizon, based on the plan computed at time-step $t$, while $\gamma_i(t+\tau|t) = 0$ implies that the facet has not yet been scheduled for inspection as of time-step $t+\tau$. 

To enable the agent to design long-horizon inspection trajectories, anticipate which facets can be inspected at future time-steps while integrating past information (i.e., facets already inspected), we design the following inspection planning dynamics:
\begin{subequations}
\begin{align} 
& \xi_i(t+\tau+1|t) - \xi_i(t+\tau|t) = \varpi_i(\tau)  \label{eq:inspection_dyn1} \\ 
& \varpi_i(\tau) \leq \gamma_i(t+\tau|t) + \Xi_i(t) \label{eq:inspection_dyn2} \\ 
& \xi_i(t|t) = \xi_i(t|t-1)  \label{eq:inspection_dyn3}
\end{align}
\end{subequations}
\noindent where the variable $\xi_i(t+\tau+1|t) \in [0,1]$ captures the discrete-time dynamics of the inspection value for each facet via the difference equation \eqref{eq:inspection_dyn1}, initialized from the previous plan, and $\varpi_i(\tau) \in [0,1]$ is an auxiliary decision variable which integrates the inspection value $\gamma_i(t+\tau|t)$ during the current planning horizon with a memory component $\Xi_i(t) \in [0,1]$. Specifically, $\Xi_i(t)=1$ only if there exists a time-step $t^\prime < t$ at which the facet $s_i$ has been inspected. Subsequently, $\xi_i(t+\tau+1|t)$ captures the evolution of inspection value for each facet based on current and past information up to time-step $t$. Based on this, the agent's objective can be re-formulated as: $\max \sum_{i=1}^{|\Delta \hat{\mathcal{S}}|} r_i \xi_i(t+N|t)$
\noindent where \( \boldsymbol{\gamma} \in \{0,1\}^{|\Delta \hat{\mathcal{S}}| \times N} \) is the collection of binary decision variables \( \gamma_i(t+\tau|t)~ \forall i, \forall \tau \). Observe that the inspection value of some facet $i$ can be activated either through \( \Xi_i(t) \) or via \( \gamma_i(t+\tau|t) \). 
Consequently, if a facet \( i \) has been inspected at some time-step \( t^\prime < t \) in the past (i.e., \( \Xi_i(t) = 1 \)), then turning on \( \gamma_i(t+\tau \mid t) \) (i.e., planning to inspect it again at some future time step \( t+\tau \mid t \)) does not improve the objective. Therefore, in such cases, the agent is encouraged to design a plan for inspecting some other facet \( j \neq i \) that has not yet been inspected, which will improve the objective. Next, we discuss the design of perception-aware constraints based on back-face elimination, enabling the agent to identify visible facets along its path and plan accurate long-horizon inspection strategies. Additionally, we explain how \( \boldsymbol{\gamma} \) is integrated with perception-aware constraints within a data-driven predictive control framework to guide the agent in maximizing the cumulative reward.

\subsection{Back-face Elimination} \label{ssec:backface_elimination}
Back-face elimination \cite{pantazopoulos2002occlusion} is an optimization technique in 3D computer graphics that enhances rendering performance by removing polygons (typically triangles) oriented away from the camera. Since these back-facing polygons are not visible to the viewer, discarding them reduces the number of polygons that need to be processed and rendered, thereby conserving computational resources. This method determines a polygon's visibility by assessing the orientation of its normal vector relative to the viewing direction.
In this work, this methodology is utilized to integrate perception-aware constraints to 3D inspection planning aimed at determining facet visibility relative to the agent's predicted state, thereby enabling the design of accurate inspection plans. More specifically, a facet is considered visible if its normal vector is oriented opposite to the camera's viewing direction (i.e., the camera is facing the facet). Additionally, the facet's centroid must lie within the convex hull of the camera's field of view, a criterion that can be extended to include all vertices of the facet.
Let the agent's camera viewing direction at a given orientation and position be denoted by \(\vec{c}_{\theta_z,\theta_y}(t+\tau|t) = c^o_{\theta_z,\theta_y} - hy(t+\tau|t)\), for $\tau \in \{0,\ldots,N-1\}$, where  \(c^o_{\theta_z,\theta_y} = \text{mean}(V_{\theta_z,\theta_y}(1), \ldots, V_{\theta_z,\theta_y}(4))\) is the centroid of the FOV's base. For a given facet \(s_i, i \in \{1,\ldots,|\Delta \hat{\mathcal{S}}|\}\) with an outward normal vector \(\vec{s}_i\)  and centroid \(\bar{s}_i\), we design the following perception-aware constraints to determine visibility:
\begin{subequations}
\begin{align} \label{eq:vis_con1}
&\text{dot}\left(\vec{c}_{\theta_z,\theta_y}(t+\tau|t),\vec{s}_i \right) \leq M\left(1-b_{i,\theta_z,\theta_y}(t+\tau|t)\right) \\ \label{eq:vis_con2}
& \text{dot}\left(\vec{c}_{\theta_z,\theta_y}(t+\tau|t),\vec{s}_i \right) > -M b_{i,\theta_z,\theta_y}(t+\tau|t) \\ \label{eq:vis_con3}
& \psi_{i,\theta_z,\theta_y}(t+\tau|t)=1 \implies \bar{s}_i \in C_{\theta_z,\theta_y}(t+\tau|t)   
\end{align}
\end{subequations}

\noindent where $\text{dot}(a,b)$ is the dot-product between vectors $a$ and $b$, \( M \) is a large positive constant, and \( b_{i,\theta_z,\theta_y}(t+\tau|t) \in \{0,1\} \), \( \psi_{i,\theta_z,\theta_y}(t+\tau|t) \in \{0,1\} \) are binary decision variables indicating whether the camera faces the facet \( s_i \) and whether its centroid \( \bar{s}_i \) belongs to the convex hull of the camera's FOV, respectively. More specifically, if \( b_{i,\theta_z,\theta_y}(t+\tau|t) = 1 \), then from Eq. \eqref{eq:vis_con1}-\eqref{eq:vis_con2}, the dot product between \( \vec{c}_{\theta_z,\theta_y}(t+\tau|t) \) and \( \vec{s}_i \) is negative, and restricted to the range \( (-M,0] \), indicating visibility, where the camera viewing direction faces the facet's normal vector. On the other hand, if the dot product between these two vectors is positive, it indicates that they are not pointing toward each other, therefore, the camera does not face the facet. Consequently, if \( \text{dot}\left(\vec{c}_{\theta_z,\theta_y}(t+\tau|t), \vec{s}_i \right) > 0 \), the decision variable \( b_{i,\theta_z,\theta_y}(t+\tau|t) \) is forced to become zero to satisfy the constraints within the interval \( (0, M] \). Finally, the binary decision variable \( \psi_{i,\theta_z,\theta_y}(t+\tau|t) \) is activated only if the centroid \( \bar{s}_i \) resides within the convex hull of the camera's FOV at time-step \( t+\tau|t \) of the planning horizon as shown in \eqref{eq:vis_con3}. We note here that the first two constraints above can be simplified since the camera direction vector \(\vec{c}_{\theta_z,\theta_y}(t+\tau|t)\) remains unchanged under translation. Therefore, the constraints in Eq. \eqref{eq:vis_con1}-\eqref{eq:vis_con2} are time-invariant, allowing the time index to be dropped from all variables (i.e., the viewing direction vector can be computed by rotating the camera at the origin).

 

\subsection{Data-Driven Predictive Inspection Control}

Given our UAS agent represented by \( \mathscr{B}(A, B, C, D) \) as described in Sec. \ref{ssec:kinematic_model}, and based on Lemma 1, we collect a sequence of input \( u^d \in \mathbb{R}^{m \times T} \) and output \( y^d \in \mathbb{R}^{p \times T} \) data of length \( T \geq (m+1)(L+n)-1 \), which are persistently exciting of order \( L+n \), with \( L \in \mathbb{N}^+ = K+N \). Here, \( K \geq \ell \) denotes the length of the initialization horizon (i.e., a window of historical input-output data) used to capture the initial state of the system at the start of the predictive control horizon, ensuring that predictions for future actions are consistent with the system's past behavior, i.e., the executed trajectory. Subsequently, \( \ell \in \mathbb{N}^+ \) denotes the smallest integer such that the rank of the observability matrix \( \mathcal{O}_\ell(A, C) = \begin{bmatrix} C^\top,(CA)^\top,\cdots,(CA^{\ell-1})^\top \end{bmatrix}^\top \in \mathbb{R}^{p \ell \times n} \) is equal to \( n \), i.e., \( \text{rank}(\mathcal{O}_\ell(A, C)) = n \) \cite{Markovsky2008data}. Finally, as already discussed, \( N \) is the planning horizon, defining the length of time-steps into the future to predict the system. Subsequently, the system's dynamical behavior is captured through the two Hankel matrices: $U_H = \mathcal{H}_{L}(u^d),~\text{and}~ Y_H = \mathcal{H}_{L}(y^d)$, where \( U_H \) is divided into two parts (i.e., \( U_p \) and \( U_f \)), used to capture the past and future behavior of the system, respectively. Here, \( U_p \in \mathbb{R}^{mK \times T-L+1} \) and \( U_f \in \mathbb{R}^{mN \times T-L+1} \). The same procedure applies to the Hankel matrix \( Y_H \) for the output data.
The proposed controller is formulated as a quadratic mixed-integer optimization problem (MIQP) shown in Problem (P1) via Eq. \eqref{eq:P1_a}-\eqref{eq:P1_u}. This controller, is initialized based on the past input-output sequence \( (u_o, y_o) \) of length \( K \), which effectively sets the underlying initial states from which the predicted trajectory \( (u, y) \) will evolve. This initialization procedure is shown in Eq. \eqref{eq:P1_b}-\eqref{eq:P1_e}, where \( u_o \in \mathbb{R}^{m \times K} \) and \( y_o \in \mathbb{R}^{p \times K} \), so that \( u_o(t) \in \mathbb{R}^{m} \) and \( y_o(t) \in \mathbb{R}^{p} \). At each time-step \( t \), the sequence pair \( (u_o, y_o) \) tracks the executed trajectory, updating its values by shifting those from the previous time-step one position to the left (i.e., Eq. \eqref{eq:P1_b}, \eqref{eq:P1_d}), and then incorporating the applied control input \( u(t-1|t-1) \) and observed output \( y(t-1|t-1) \) from the previous time-step \( t-1 \) (i.e., Eq. \eqref{eq:P1_c}, \eqref{eq:P1_e}). 
\begin{algorithm}
\begin{subequations}
\begin{align} 
&\hspace*{-3mm}\textbf{(P1)}~\textbf{Predictive Planning and Control} & \notag\\
&\hspace*{-3mm}~~~~~~\underset{\boldsymbol{u}}{\text{min}} ~\mathcal{J} &   \label{eq:P1_a} \\
&\hspace*{-3mm}\textbf{subject to: } ~  &\nonumber\\
&\hspace*{-3mm}\texttt{- Initialize past behavior:} ~  &\nonumber\\
&\hspace*{-3mm} u_o(t-\kappa|t) = u_o(t-\kappa+1|t-1), & \hspace*{-25mm} \kappa \in \{1,..,K-1\} \label{eq:P1_b}\\
&\hspace*{-3mm} u_o(t-K|t) = u(t-1|t-1) &  \label{eq:P1_c}\\
&\hspace*{-3mm} y_o(t-\kappa|t) = y_o(t-\kappa+1|t-1), & \hspace*{-25mm} \kappa \in \{1,..,K-1\} \label{eq:P1_d}\\
&\hspace*{-3mm} y_o(t-K|t) = y(t-1|t-1) & \label{eq:P1_e} \\
&\hspace*{-3mm}\texttt{- Construct future behavior:} ~  &\hspace*{-25mm} \nonumber\\
&\hspace*{-3mm} \text{dot}\left(U_p(\kappa),g\right) = u_o(t-\kappa|t),  & \hspace*{-25mm} \forall \kappa \label{eq:P1_f}\\
&\hspace*{-3mm} \text{dot}\left(Y_p(\kappa),g\right) = y_o(t-\kappa|t), & \hspace*{-25mm} \forall \kappa  \label{eq:P1_g}\\
&\hspace*{-3mm} u(t+\tau|t) = \text{dot}\left(U_f(\tau+1),g\right), & \hspace*{-25mm} \forall \tau \label{eq:P1_h}\\
&\hspace*{-3mm} y(t+\tau|t) = \text{dot}\left(Y_f(\tau+1),g\right), & \hspace*{-25mm} \forall \tau \label{eq:P1_i}\\
&\hspace*{-3mm}\texttt{- Inspection planning dynamics:} ~  &\hspace*{-25mm} \nonumber\\
&\hspace*{-3mm}\xi_i(t+\tau+1|t) = \xi_i(t+\tau|t) + \varpi_i(\tau), & \hspace*{-25mm} \forall \tau,i \label{eq:P1_j}\\
&\hspace*{-3mm}\xi_i(t|t) = \xi_i(t|t-1) , & \hspace*{-25mm} \forall i \label{eq:P1_k}\\
&\hspace*{-3mm} \varpi_i(\tau) \leq \gamma_i(t+\tau|t) + \Xi_i(t) , & \hspace*{-25mm} \forall i \label{eq:P1_l}\\
&\hspace*{-3mm}\texttt{- Perception-aware constraints:} ~  &\hspace*{-25mm} \nonumber\\
&\hspace*{-3mm} \text{dot}\left(\vec{c}_{\phi},\vec{s}_i \right) \leq M\left(1-b_{i,\phi}\right), & \hspace*{-25mm} \forall i,\phi \label{eq:P1_m}\\
&\hspace*{-3mm} \text{dot}\left(\vec{c}_{\phi},\vec{s}_i \right) > -M b_{i,\phi}, & \hspace*{-25mm} \forall i,\phi \label{eq:P1_n}\\
&\hspace*{-3mm} \psi_{i,\phi}(t+\tau|t)=1 \implies & \hspace*{-25mm} \notag \\
&\hspace*{-3mm} ~~~~~~~~~~~~~~~~~~\bar{s}_i \in C_{\phi}(t+\tau|t) , & \hspace*{-25mm} \forall i,\phi \label{eq:P1_o}\\
&\hspace*{-3mm}\texttt{- Constraint Integration:} ~  &\hspace*{-25mm} \nonumber\\
&\hspace*{-3mm} \gamma_i(t+\tau|t) \leq \left[~b_{i,\phi} \times \psi_{i,\phi}(t+\tau|t)~\right] \times & \hspace*{-25mm} \notag \\
&\hspace*{-3mm} ~~~~~~~~~~~~~~~~~\omega_{\phi}(t+\tau|t), & \hspace*{-25mm} \forall \tau,i,\phi \label{eq:P1_p}\\
&\hspace*{-3mm} \texttt{- Collision avoidance:} ~  &\hspace*{-25mm} \nonumber\\
&\hspace*{-3mm} y(t+\tau|t) \notin C_{\Delta \mathcal{S}} & \hspace*{-25mm} \forall \tau \label{eq:P1_q}\\
&\hspace*{-3mm} \texttt{- Variable definitions:} ~  &\hspace*{-25mm} \nonumber\\
&\hspace*{-3mm} \tau \in \{0,..,N-1\}, \kappa \in \{1,..,K\}, i \in \{1,..,|\Delta \hat{\mathcal{S}}|\} & \hspace*{-25mm} \label{eq:P1_r} \\
&\hspace*{-3mm} \phi \in \{1,..,|\{\Theta_z \times \Theta_y\}|\}, \omega_{\phi}(t+\tau|t) \in \{0,1\} & \hspace*{-25mm} \label{eq:P1_s}\\
&\hspace*{-3mm} \gamma_i(t+\tau|t),\psi_{i,\phi}(t+\tau|t),b_{i,\phi} \in \{0,1\} , & \hspace*{-25mm} \label{eq:P1_t}\\
&\hspace*{-3mm} \xi_i(t+\tau+1|t),\Xi_i(t),\varpi_i(\tau) \in [0,1], g \in \mathbb{R}^{T-L+1}& \hspace*{-25mm} \label{eq:P1_u}
\end{align}
\end{subequations}
\vspace{-5mm}
\end{algorithm}

The constraints shown in Eq. \eqref{eq:P1_f}-\eqref{eq:P1_i}, construct the future trajectory of the agent by identifying a consistent future input-output sequence $(u,y)$ of length $N$ aligned  with the past data $(u_o,y_o)$ of length $K$, via the decision variable $g \in \mathbb{R}^{T-L+1}$. The notation \( U_p(\kappa) \) in Eq. \eqref{eq:P1_f} represents the \( \kappa_\text{th} \) block row of \( U_p \in \mathbb{R}^{mK \times T-L+1} \), so that \( U_p(\kappa) \in \mathbb{R}^{m \times T-L+1} \). Subsequently, the dot product operation \( \text{dot}(U_p(\kappa), g) = u_o(t-\kappa|t) \) is performed between all \( m \) rows of \( U_p(\kappa) \) and \( g \), resulting in an \( m \)-dimensional signal, which we equate to \( u_o(t-\kappa|t) \). The same notation is applied to the remaining constraints in Eq. \eqref{eq:P1_g}-\eqref{eq:P1_i}. The predicted control inputs and system outputs over the rolling planning horizon \( \tau \in \{0, \ldots, N-1\} \) are represented by the decision variables \( u(t+\tau|t) \) and \( y(t+\tau|t) \).

The constraints in Eq. \eqref{eq:P1_j}-\eqref{eq:P1_l} capture the inspection planning dynamics discussed in Sec. \ref{ssec:inspection_dynamics}. These constraints are designed to minimize redundant work by (a) tracking already inspected facets through the memory component \( \Xi_i(t) \) and (b) preventing the generation of plans that inspect the same facet more than once within the planning horizon. 

Next, the constraints in Eq. \eqref{eq:P1_m}-\eqref{eq:P1_n} identify the visible facets through back-face elimination as discussed in Sec. \ref{ssec:backface_elimination}, where with slight change of notation $\phi \in \{1,\ldots,|\{\Theta_z \times \Theta_y\}|\}$ serves as index pointing to a specific FOV orientation $\{\theta_z,\theta_y\} \in \{\Theta_z \times \Theta_y\}$. The viewing direction \( \vec{c}_\phi, \forall \phi \), is precomputed for all camera FOV rotations since it is invariant under translation. Therefore, the binary decision variable \( b_{i,\phi} \) indicates whether facet \( i \) can be observed through the camera orientation \( \phi \), based on the facet's normal vector and the camera's viewing direction.
To determine visibility, this result must be combined with the binary decision variable \( \psi_{i,\phi}(t+\tau|t) \), which indicates whether the centroid of facet \( s_i \) resides within the convex hull of the FOV with orientation \( \phi \) at time step \( t+\tau|t \). 
 This is achieved with the constraint shown in Eq. \eqref{eq:P1_o} as follows: The convex hull of the agent's \( \phi_\text{th} \) camera FOV at time step \( t+\tau|t \) is given, as discussed in Sec. \ref{ssec:cam_model}, by \( C_{\phi}(t+\tau|t) = \{\text{x} \in \mathbb{R}^3 \mid \Gamma_{\phi} \text{x} \leq \Delta'_{\phi}(t+\tau|t)\} \), where \( \Delta'_{\phi}(t+\tau|t) = \Delta_{\phi} + \Gamma_{\phi} hy(t+\tau|t) \) and \( hy(t+\tau|t) \) indicates the predicted output of the system, i.e., the agent's position inside the planning horizon. 
 The binary decision variable $\psi_{i,\phi}(t+\tau|t)$ can then be derived from the auxiliary binary variable $\psi^\prime_{j,i,\phi}(\tau)$ via the constraint:
\begin{align}
	\text{dot}\left(\Gamma_\phi(j),\bar{s}_i\right) &+ M \psi^\prime_{j,i,\phi}(\tau)- \Delta_{\phi}(j) \psi^\prime_{j,i,\phi}(\tau)~ -  \notag\\
	&  \text{dot}\left(\Gamma_{\phi}(j), hy(\tau)\right)\psi^\prime_{j,i,\phi}(\tau) \leq M, \label{eq:f1}
\end{align}

\noindent where for brevity $t+\tau|t$ is abbreviated as $\tau$, $j \in \{1,\ldots,5\}$ denotes the FOV faces, $M$ is a large positive constant, $\Gamma_\phi(j)$ denotes the $j_\text{th}$ row of the matrix $\Gamma_\phi$, $\Delta_{\phi}(j)$ is the $j_\text{th}$ element of the column vector  $\Delta_{\phi}$, and $\psi^\prime_{j,i,\phi}(\tau) \in \{0,1\}$. $\psi^\prime_{j,i,\phi}(\tau)$ will take the value of 1 if the inequality \( \text{dot}(\Gamma_\phi(j),\bar{s}_i) \leq \Delta_{\phi}(j) + \text{dot}(\Gamma_{\phi}(j), hy(\tau)) \) holds for some \( j, \phi, i \), and \( \tau \); otherwise, it will take the value of 0. Subsequently, $\bar{s}_i \in  C_{\phi}(t+\tau|t)$ implies that $\psi^\prime_{j,i,\phi}(\tau)=1, \forall j$. This can be implemented with the following constraint: $5\psi_{i,\phi}(\tau)  \leq \sum_{j=1}^5 \psi^\prime_{j,i,\phi}(\tau), \forall \phi,i,\tau$, 
\noindent where the controllable binary decision variable \( \psi_{i,\phi}(\tau) \), when driven to 1 forces  $\psi^\prime_{j,i,\phi}(\tau)=1, \forall j$ in order to satisfy the constraint. In essence, by setting \( \psi_{i,\phi}(t+\tau|t) \) to 1, we are effectively guiding the agent's \( \phi_\text{th} \) camera FOV through the output \( hy(t+\tau|t) \) at time step \( t+\tau|t \) to cover the centroid of facet \( s_i \).

Next, the perception-aware constraints are given in Eq. \eqref{eq:P1_p} via the decision variable \( \gamma_i(t+\tau|t) \), which indicates that there exists a camera FOV orientation such that facet \( s_i \) is visible and can be inspected via the agent's output at time-step \( t+\tau|t \). This is accomplished by multiplying the binary decision variables \( b_{i,\phi} \) (back-face elimination), \( \psi_{i,\phi}(t+\tau|t) \) (centroid inside the FOV), and \( \omega_\phi(t+\tau|t) \in \{0,1\} \). In particular $\omega_\phi(t+\tau|t)$ ensures that, at each time-step within the planning horizon, the camera orientation is restricted to one active configuration, i.e., \( \sum_\phi \omega_\phi(t+\tau|t) = 1, \forall \tau \). This multiplication, can easily be implemented in modern mixed-integer solvers using the logical AND operator. To ensure the UAS agent avoids collision with the structure \( \mathcal{S} \), we enforce the constraint in Eq. \eqref{eq:P1_q}. The convex hull of \( \mathcal{S} \) is defined by the intersection of \( |\Delta\mathcal{S}| \) half-spaces. Each half-space \( j \), for \( j = \{1, \ldots, |\Delta\mathcal{S}| \}\), corresponds to a plane described by the equation \( \text{dot}(\alpha(j), \text{x}) = \beta(j) \) with \( \text{x} \in \mathbb{R}^3 \), effectively partitioning the 3D space into two regions. To prevent a collision with \( \mathcal{S} \), the following conditions must be satisfied for all time-steps \( \tau \) and all planes \( j \):
\begin{equation}
	\text{dot}\left(\alpha(j), hy(t+\tau|t)\right) > \beta(j) - M o_{j}(t+\tau|t),~\forall \tau, j, \label{eq:O_1}
\end{equation}
%
%
\noindent where \( M \) is a sufficiently large positive constant, and \( o_{j}(t+\tau|t) \in \{0,1\} \) are binary variables. When \( o_{j}(t+\tau|t) = 1 \), it indicates that the inequality \( \text{dot}(\alpha(j), hy(t+\tau|t)) \leq \beta(j) \) holds at time-step \( t+\tau|t \), suggesting potential collision along plane \( j \). The agent is considered to be in collision with \( \mathcal{S} \) at time-step \( t+\tau|t \) - meaning it resides within the convex hull defined by \( \Delta\mathcal{S} \) if \( \text{dot}(\alpha(j), hy(t+\tau|t)) \leq \beta(j) \) holds for all \( j \). 
Subsequently, the constraint $\sum_{j=1}^{|\Delta\mathcal{S}|} o_{j}(t+\tau|t) + 1 \le |\Delta\mathcal{S}|, ~ \forall \tau$ ensures that the total number of activated \( o_{j}(t+\tau|t) \) variables at any given time-step is less than \( |\Delta\mathcal{S}| \), thereby preventing the agent from simultaneously satisfying all the inequalities that define the convex hull of \( \mathcal{S} \) and thus avoiding collision.
Finally, the cost function $\mathcal{J}$ to be minimized is given by:
\begin{align}\label{eq:main_objective}
	\mathcal{J} &= w_1\sum_{\tau=1}^{N-1} ||\Delta u(t+\tau|t)||^2_2 -  w_2\sum_{i=1}^{|\Delta \hat{\mathcal{S}}|} r_i \xi_i(t+N|t)
\end{align}
	\noindent where $\Delta u(t+\tau|t) = u(t+\tau|t) - u(t+\tau-1|t), \tau \in \{1,\ldots,N-1\}$, and the parameters \( w_1 \) and \( w_2 \) are tuning weights used to balance control effort and inspection performance, respectively.
	

\begin{figure*}
	\centering
	\includegraphics[width=\textwidth]{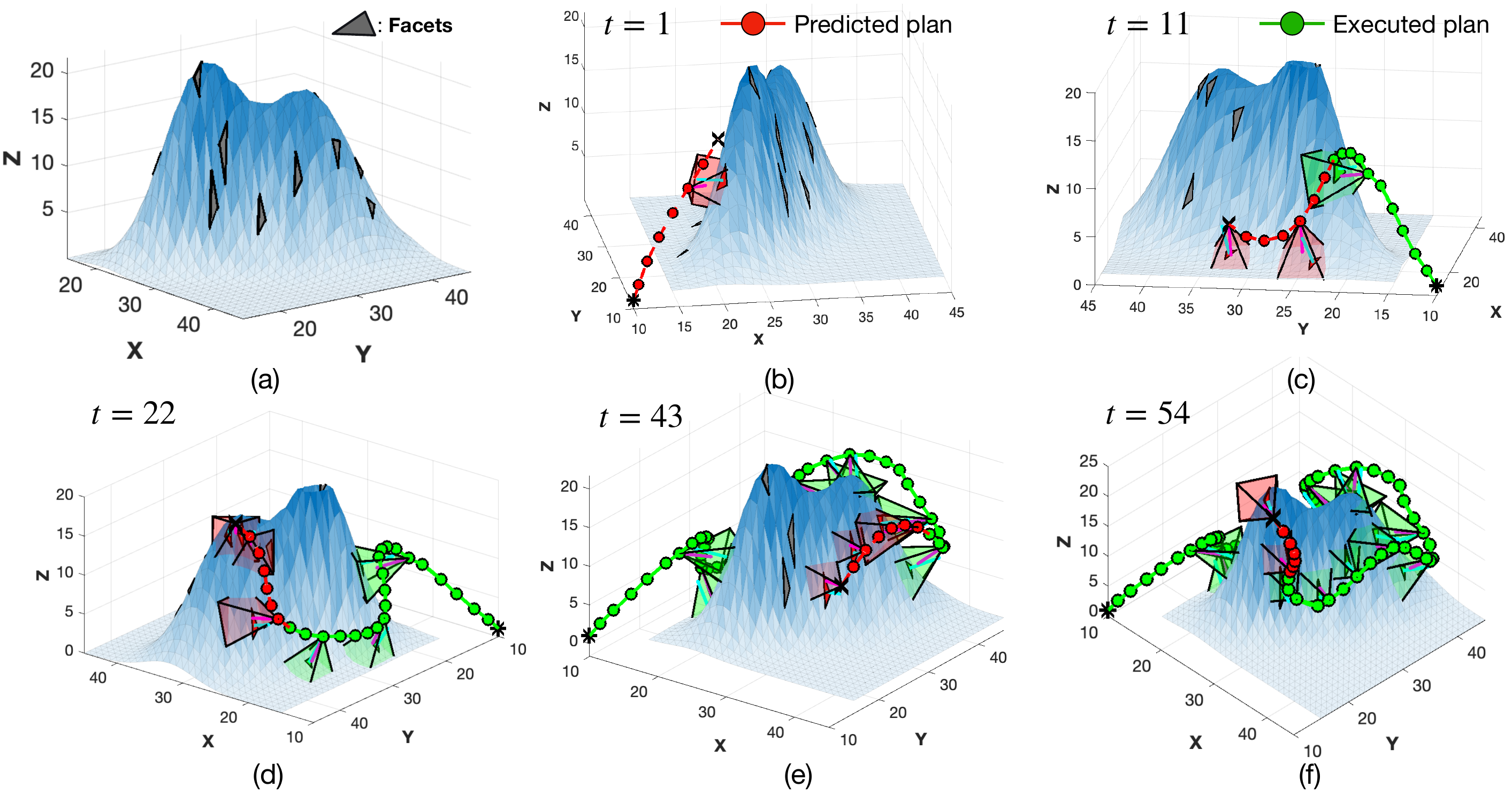}
	\caption{The figure provides an example to illustrate the proposed approach of data-driven predictive planning and control for 3D inspection.}
	\label{fig:res1}
	\vspace{-0mm}
\end{figure*}

\section{Evaluation} \label{sec:Evaluation}
\subsection{Simulation Setup}\label{ssec:setup}

To evaluate our approach, we simulated a quadrotor UAS agent with a 12-dimensional state vector \( x(t) = [x_1, \dot{x}_1, x_2, \dot{x}_2, x_3, \dot{x}_3, x_4, \dot{x}_4, x_5, \dot{x}_5, x_6, \dot{x}_6]^\top \), and full state measurement \( y(t) \). The state vector includes both 3D Cartesian position and velocity components, \( (x_1, \dot{x}_1) \), \( (x_2, \dot{x}_2) \), \( (x_3, \dot{x}_3) \), corresponding to the \( x \)-, \( y \)-, and \( z \)-axes, as well as angular positions and velocities, \( (x_4, \dot{x}_4) \), \( (x_5, \dot{x}_5) \), and \( (x_6, \dot{x}_6) \), representing roll, pitch, and yaw, respectively.

Quad-rotor control \( u(t) = [u_1, u_2, u_3, u_4]^\top \) is achieved by managing four independent inputs, where the lift force \( u_1 \in [-3,3] \)N along the z-axis controls vertical motion. Additionally, torques \( u_2 \), \( u_3 \), and \( u_4 \) are applied around the \( x \)-, \( y \)-, and \( z \)-axes, respectively, to control roll, pitch, and yaw, with $u_2, u_3, u_4 \in [-2,2]$Nm.

The dynamical behavior of the UAS is linearized around the hover state and is described by \( \mathcal{B}(A, B, C, D) \) with \( A \in \mathbb{R}^{12 \times 12} \), \( B \in \mathbb{R}^{12 \times 4} \), \( C = I_{12} \) (identity matrix), and \( D = 0 \). The non-zero elements of \( A \) and \( B \) are as follows: \( A(1,2) = T_s \), \( A(2,2) = 1 \), \( A(2,9) = g_r T_s \), \( A(3,4) = T_s \), \( A(4,4) = 1 \), \( A(4,7) = -g_r T_s \), \( A(5,6) = T_s \), \( A(6,6) = 1 \), \( A(7,8) = T_s \), \( A(8,8) = 1 \), \( A(9,10) = T_s \), \( A(10,10) = 1 \), \( A(11,12) = T_s \), \( A(12,12) = 1 \), and \( B(6,1) = \frac{T_s}{ma} \), \( B(8,2) = \frac{l T_s}{I_x} \), \( B(10,3) = \frac{l T_s}{I_y} \), \( B(12,4) = \frac{T_s}{I_z} \).

The parameters \( T_s = 0.7 \, \text{s} \), \( g_r = 9.81 \, \text{m/s}^2 \), \( ma = 1.2 \, \text{kg} \), \( l = 0.21 \, \text{m} \), and \( I_x = I_y = I_z = 0.004 \, \text{N} \cdot \text{s}^2/\text{rad} \) denote the sampling time, gravity, UAV mass, quadrotor arm length, and moments of inertia about the $x-$, $y-$, and $z-$axes, respectively. The camera FOV at the origin is given by $
V_\text{FOV} = \begin{bmatrix}
\begin{smallmatrix}
 -W/2 & W/2 & W/2 & -W/2 & 0 \\
 W/2 & W/2 & -W/2 & -W/2 & 0 \\
 H & H & H & H & 0 \\
\end{smallmatrix}
\end{bmatrix}
$ with \( W = 5.5 \, \text{m} \) and \( H = 6.5 \, \text{m} \). The camera can perform \( |\Theta_z| = |\Theta_y| = 8 \) equally spaced rotations around the \( y \)- and \( z \)-axes. The parameters $w_1$ and $w_2$ in Eq. \eqref{eq:main_objective} are set to $5E-3$ and 15 respectively, and the controller in Problem (P1) is implemented utilizing the Gurobi MIQP solver.

\vspace{-1mm}
\subsection{Results}

Figure \ref{fig:res1} illustrates a 3D inspection scenario executed using the proposed controller from Problem (P1), configured with parameters \( K = 1 \), \( N = 8 \), and a  collected input-output data sequence of length $T=(m+1)(L+n)-1 = 104$, where $L=K+N$. The structure $\mathcal{S}$ to be inspected is represented as a mesh $\Delta \mathcal{S}$ ($|\Delta \mathcal{S}|=1008$)  shown in Fig. \ref{fig:res1}(a), with the facets to be inspected $\Delta \hat{\mathcal{S}}$ shown in gray color, randomly sampled from $\Delta \mathcal{S}$, with cardinality $|\Delta \hat{\mathcal{S}}|=15$, and assigned rewards $r_i, i \in \{1,\ldots,15\}$ randomly sampled within the interval $[1,20]$. The agent is initialized at the positional state of \( (x, y, z) = (10, 10, 1) \), marked with $\ast$ as shown in Fig. \ref{fig:res1}(b). The evolution of the inspection mission is shown at different time-steps in Fig.~\ref{fig:res1}(b)-\ref{fig:res1}(f), showcasing the predicted and executed plan (i.e., trajectory and camera FOV) in red and green, respectively. As shown in the figure, the approach simultaneously solves planning and control under perception-aware constraints. The agent determines in an online manner the sequence of facets to inspect next, by optimizing the cumulative reward while considering visibility, within the planning horizon, and generates the optimal trajectory and camera states accordingly. The mission concludes after 60 time-steps, once all facets have been inspected.

 \begin{figure}
	\centering
	\includegraphics[width=\columnwidth]{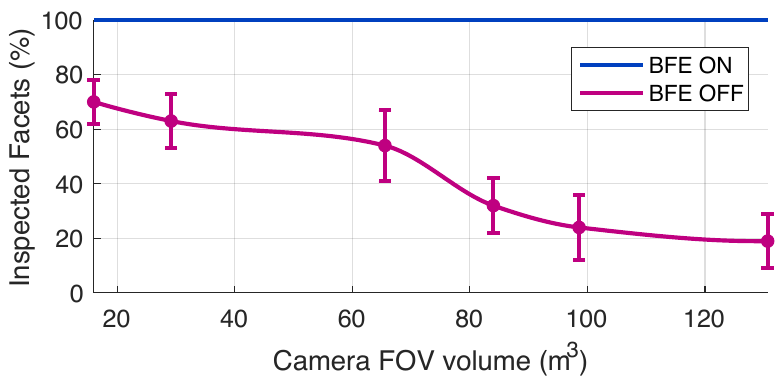}
	\caption{The effect of back-face elimination (BFE) on the performance of 3D inspection planning. Average percentage of visible inspected facets at the end of the mission as a function of the FOV size.}
	\label{fig:fig2}
	\vspace{-5mm}
\end{figure}

Finally, the next experiment evaluates the significance of back-face elimination (BFE) in identifying visible facets for the inspection task. A Monte Carlo simulation with 50 trials was conducted, randomly initializing a UAS agent in the simulated environment shown in Fig. \ref{fig:res1}, and running the proposed controller with and without BFE constraints (i.e., by setting $b_{i,\phi}=1,  \forall i,\phi$ in Eq. \eqref{eq:P1_p}). The facets for inspection were randomly selected from $\Delta \mathcal{S}$, with their quantity randomly sampled from the range [10, 30]. Ground truth was obtained through ray-tracing \cite{10925885}. The findings, illustrated in Fig. \ref{fig:fig2}, reveal that with BFE enabled, the controller successfully inspects all facets regardless of FOV size, without performance loss. However, disabling BFE results in a performance drop as FOV size increases. This is due to larger FOVs including facets that are within the convex hull of the FOV but not truly visible. Small FOVs require the agent to move closer to the structure, naturally aligning with visible facets. In contrast, larger FOVs may cover more facets in the structure, even if some are not actually visible. Hence, the incorporation of back-face elimination (BFE) enables perception-aware guidance, thereby enhancing the effectiveness of the inspection process.

\section{Conclusion} \label{sec:conclusion}

This paper presents a novel UAS-based, data-driven control approach to 3D inspection planning. By integrating perception, planning, and control within a unified predictive control framework, the method dynamically generates optimal inspection trajectories guided by a perception-aware objective aimed at maximizing inspection performance. The inspection planning task is formulated as an optimization problem that maximizes cumulative rewards over a receding planning horizon, incorporating back-face elimination into the control loop for online visibility determination. Furthermore, this work explores the application of data-driven control to complex optimization problems beyond traditional regulation and tracking tasks. Future work will focus on deploying the proposed approach in real-world scenarios using off-the-shelf UAS platforms.

\flushbottom
\balance

\bibliographystyle{IEEEtran}
\bibliography{IEEEabrv,main}

\end{document}